\pdfoutput=1

\documentclass[11pt]{article}

\usepackage[preprint]{acl}

\usepackage{times}
\usepackage{latexsym}
\usepackage{amsthm}
\usepackage[english]{babel}

\usepackage[T1]{fontenc}

\usepackage[utf8]{inputenc}

\usepackage{microtype}

\usepackage{inconsolata}

\usepackage{graphicx}

\usepackage{enumitem}
\usepackage{amsmath}

\usepackage{colortbl} 
\usepackage{xcolor} 
\usepackage[most]{tcolorbox}
\usepackage{booktabs,multirow}

\usepackage{siunitx}

\definecolor{c1}{cmyk}{0,0.6175,0.8848,0.1490} 
\definecolor{c2}{cmyk}{0.1127,0.6690,0,0.4431} 
\definecolor{c3}{cmyk}{0.3081,0,0.7209,0.3255} 
\definecolor{c4}{cmyk}{0.6765,0.2017,0,0.0667} 
\definecolor{c5}{cmyk}{0,0.8765,0.7099,0.3647} 

\newtcbox{\hlprimarytab}{on line, rounded corners, box align=base, colback=c3!10,colframe=white,size=fbox,arc=3pt, before upper=\strut, top=-2pt, bottom=-4pt, left=-2pt, right=-2pt, boxrule=0pt}
\newtcbox{\hlsecondarytab}{on line, box align=base, colback=red!10,colframe=white,size=fbox,arc=3pt, before upper=\strut, top=-2pt, bottom=-4pt, left=-2pt, right=-2pt, boxrule=0pt}

\newcommand{\dashifted}{\raisebox{0.5ex}{\tiny$\downarrow$}}
\newcommand{\uashifted}{\raisebox{0.5ex}{\tiny$\uparrow$}}
\newcommand{\da}[1]{{~\scriptsize\hlprimarytab{\dashifted{#1}}}}
\newcommand{\ua}[1]{{~\scriptsize\hlsecondarytab{\uashifted{#1}}}}
\newcommand{\uag}[1]{~{\scriptsize\hlprimarytab{\uashifted{#1}}}}
\newcommand{\dab}[1]{~{\scriptsize\hlsecondarytab{\dashifted{#1}}}}

\title{
Gender Encoding Patterns in Pretrained Language Model Representations
}

\author{ \textbf{Mahdi Zakizadeh} \\ 
TeIAS, 
Khatam University, Iran \\
\texttt{m.zakizadeh@khatam.ac.ir} \\ \And
\textbf{Mohammad Taher Pilehvar} \\
Cardiff University, UK \\
\texttt{pilehvarmt@cardiff.ac.uk} \\ }

\begin{document}
\maketitle

\begin{abstract}
Gender bias in pretrained language models (PLMs) poses significant social and ethical challenges. Despite growing awareness, there is a lack of comprehensive investigation into how different models internally represent and propagate such biases. This study adopts an information-theoretic approach to analyze how gender biases are encoded within various encoder-based architectures.
We focus on three key aspects: identifying how models encode gender information and biases, examining the impact of bias mitigation techniques and fine-tuning on the encoded biases and their effectiveness, and exploring how model design differences influence the encoding of biases.
Through rigorous and systematic investigation, our findings reveal a consistent pattern of gender encoding across diverse models. 
Surprisingly, debiasing techniques often exhibit limited efficacy, sometimes inadvertently increasing the encoded bias in internal representations while reducing bias in model output distributions. This highlights a disconnect between mitigating bias in output distributions and addressing its internal representations. 
This work provides valuable guidance for advancing bias mitigation strategies and fostering the development of more equitable language models.
\footnote{The code utilized in this study is available at \url{https://github.com/mzakizadeh/Gender-Encoding-Patterns}
}

\end{abstract}

\section{Introduction}      

Pretrained language models (PLMs) have revolutionized natural language processing (NLP) by enabling a wide range of applications \cite{devlin-etal-2019-bert, Radford2019LanguageMA, Brown2020LanguageMA, Touvron2023Llama2O, Jiang2023Mistral7, Dubey2024TheL3}. These models, trained on vast amounts of data, capture intricate patterns and knowledge, including gender-related information. However, alongside their impressive capabilities, PLMs also encode harmful biases that raise significant ethical concerns \cite{silva-etal-2021-towards, field-etal-2021-survey, DBLP:journals/corr/abs-2304-07683}. These biases can perpetuate stereotypes, misrepresent individuals and groups, and lead to unfair treatment in various applications, thereby impacting social justice and equity (e.g. \citealp{park-etal-2018-reducing, kiritchenko-mohammad-2018-examining, chen2024causally, levy-etal-2024-gender}).

Understanding how PLMs encode and propagate gender information is critical for developing effective bias mitigation strategies. This challenge grows increasingly urgent with the widespread adoption of retrieval-augmented generation (RAG) techniques, which rely on encoder-derived representations to retrieve contextually relevant documents \cite{wu-etal-2025-rag}. If gender biases are deeply embedded in these encoder-derived representations, RAG pipelines risk amplifying societal biases at an unprecedented scale by retrieving and propagating stereotypical or discriminatory content.

Despite extensive research on bias in language models, much of the focus has been on identifying and measuring bias rather than comprehensively analyzing how it is embedded within the model's internal representations. Previous studies have explored bias in transformer-based models, developing metrics to quantify bias \cite{DBLP:journals/corr/IslamBN16, may-etal-2019-measuring, nangia-etal-2020-crows, nadeem-etal-2021-stereoset, felkner-etal-2023-winoqueer}, implementing techniques to reduce it \cite{zhao-etal-2018-gender, lauscher-etal-2021-sustainable-modular, kaneko-bollegala-2021-debiasing, webster-measuring, schick-etal-2021-self}, and investigating its underlying causes \cite{DBLP:conf/nips/BolukbasiCZSK16, kaneko-etal-2022-gender}. However, there remains a limited understanding of the mechanisms through which biases are encoded and how different training and fine-tuning processes influence these biases within model weights.

To address this gap, we use an information-theoretic approach, specifically Minimum Description Length (MDL) probing proposed by \citet{voita-titov-2020-information}, to explore how gender bias is encoded in various encoder-based architectures. By examining different layers of PLMs, we identify where biases emerge and how fine-tuning and debiasing techniques impact these representations. 

Our work is inspired by \citet{mendelson-belinkov-2021-debiasing} who studied the impact of debiasing techniques used to reduce the model's reliance on spurious correlations between data and labels in natural language inference on model's representations. In summary, our contributions are twofold:

\begin{itemize}[leftmargin=*] 
\item We pinpoint the specific parts of encoder-based PLMs responsible for encoding gender information, highlighting critical layers where bias is most pronounced. 
\item We assess the effect of various debiasing methods, demonstrating that pretrained debiasing objectives outperform post-hoc mitigation approaches in reducing encoded bias. 
\end{itemize}

\section{Related Works}

In this section, we review some of the related studies on gender bias in language models, bias mitigation and measurement methods, and probing techniques and their use in bias evaluation.

\subsection{Bias in Language Models}
Early investigations into gender bias in language models unveiled that static embeddings not only encode but also amplify human-like biases within their representations \cite{DBLP:journals/corr/IslamBN16, DBLP:conf/nips/BolukbasiCZSK16}. Subsequently, various studies have proposed methods to manipulate the embedding space or learning algorithms to mitigate bias in such models \cite{DBLP:conf/nips/BolukbasiCZSK16, zhao-etal-2018-learning}. However, as \citet{gonen-goldberg-2019-lipstick} demonstrated, these techniques only provide superficial solutions, as biased information is not entirely removed from the model's embedding space.

The introduction of contextualized word embeddings, such as BERT \cite{devlin-etal-2019-bert}, posed new challenges, as manipulating representation space became more intricate compared to static embeddings. Contextualized language models have been shown to exhibit bias against demographic groups, including gender \cite{zhao-etal-2019-gender, silva-etal-2021-towards}. 

Despite these advancements, a comprehensive comparative analysis between various bias mitigation methods remained lacking. This gap was addressed by \citet{meade-etal-2022-empirical}, who conducted an empirical investigation into the effectiveness of multiple debiasing techniques. Through their experimentation, they selected diverse debiasing approaches, continued pretraining models with these techniques, and demonstrated their efficacy using prominent bias mitigation metrics. Additionally, they assessed the impact of these techniques on downstream performance, measuring model performance on the General Language Understanding Evaluation (GLUE; \citealp{DBLP:conf/iclr/WangSMHLB19}) test set. As the results indicated that the debiasing techniques did not significantly compromise downstream performance, they hypothesized that these methods might not negatively affect model representations. However, they did not provide concrete evidence to support their claims. This highlights the need for further research and analysis to thoroughly understand the implications and effectiveness of different debiasing techniques in the context of language models.

While earlier studies have explored the presence of gender bias in static and contextualized embeddings, they primarily focused on identifying and quantifying bias or testing basic mitigation strategies. Our study takes a different approach by investigating how biases are encoded within the internal representations of language models. This deeper exploration helps uncover where and how bias manifests, providing insights into mitigating these issues more effectively.

\subsection{Probing Techniques and Bias Evaluation}
Probing is a valuable technique for determining the knowledge characteristics captured by language models. With advancements in methods for interpreting model behavior, probing has gained traction in the research community. The introduction of Minimum Description Length probing (MDL probing; \citealp{voita-titov-2020-information}), has enabled researchers to explore the knowledge encoded in language model representations in more depth. MDL probing has been utilized to assess biases in model representations, as demonstrated by \citet{mendelson-belinkov-2021-debiasing} and \citet{orgad-etal-2022-gender}.

Intriguingly, \citet{mendelson-belinkov-2021-debiasing} found that debiasing methods intended to make models robust against spurious correlations in datasets, inadvertently led to an increase in biased information in model representations. On the other hand, \citet{orgad-etal-2022-gender} employed MDL as a metric for assessing bias and demonstrated its stronger correlation with extrinsic bias metrics used in conjunction with extrinsic bias mitigation techniques compared to other intrinsic bias measurement methods.

Building on the advancements of probing techniques, particularly the use of structured methods to interpret model behaviors, our work delves into the mechanisms by which gender biases are encoded. By systematically evaluating model layers, we aim to understand how different mitigation and fine-tuning strategies influence the internal representations of bias, extending the applications of probing techniques to new depths.

\subsection{Knowledge Localization and Bias}

Knowledge localization has emerged as a critical area of study in NLP, focusing on identifying subnets within language models that are responsible for specific tasks, domains, or linguistic properties \cite{hendy-etal-2022-domain, DBLP:conf/icml/PanigrahiSZA23, song-etal-2024-large, choenni-etal-2023-cross}. These techniques have been extended to explore gender bias, pinpointing the internal components of models that encode bias.

For example, \citet{chintam-etal-2023-identifying} employed causal inference methods, including techniques such as causal mediation analysis and differential masking, to identify attention heads responsible for biased behaviors in transformer models. Their work highlighted the ability to localize gender bias and proposed parameter-efficient fine-tuning strategies to mitigate it.
Similarly, \citet{lutz-etal-2024-local} introduced local contrastive editing, a technique leveraging unstructured pruning to precisely localize individual model weights responsible for encoding gender stereotypes. This method enabled them to edit these weights efficiently, mitigating bias without significant degradation of model performance.

Although our research aligns with prior efforts in localizing bias within pretrained language models, we introduce a distinct methodological perspective. Furthermore, by broadening the scope of experimentation across diverse models and mitigation strategies, we aim to comprehensively explore how and where gender bias is encoded. Our analysis reinforces previous findings about bias concentration in specific model layers, while also paving the way for targeted and efficient intervention techniques.

\section{Background}

Probing datasets are typically defined as $D = \{ X, Y_p \}$, where $X$ represents the input data, and $Y_p$ represents the linguistic property or knowledge we are seeking to extract from the language model. The usage of language models involves two distinct stages. In the first stage, the language model, denoted as $f_\theta: X \to Z$, transforms the input $X$ into a latent space $Z$, where $X$ denotes the textual input, $Z$ represents the latent representation of the text, and $\theta$ encompasses the model's weights. This latent space captures complex linguistic features and representations that encode the underlying information within the input text. Subsequently, in the second stage, a classifier, denoted as $g_\sigma: Z \to Y$, is employed to map the latent space $Z$ to the corresponding label space $Y$. The classifier is denoted by $g_\sigma$, with $\sigma$ encompassing its parameters. This two-stage approach facilitates the language model's ability to learn intricate language structures and encode relevant knowledge, while the classifier enables the extraction and utilization of this knowledge for various downstream tasks and analyses.

Traditionally, probing classifiers attempted to train on frozen language model weights, ensuring that the transformation from $X$ to $Z$ remains unchanged during training. Subsequently, the classifier learns how to map the latent space $Z$ to the target property space $Y_p$. If the classifier can effortlessly learn this transformation with a limited amount of data, it was concluded that the language model possesses the relevant linguistic information \cite{belinkov-2022-probing}. However, such traditional probing approaches have been shown to exhibit limitations. These methods can yield unreliable results as they tend to classify representations of random data similarly to those of actual data, indicating their inadequacy in capturing variations in representations \cite{zhang-bowman-2018-language}. As a consequence, the outcomes of these traditional probing methods are highly dependent on hyperparameter choices and might not reliably reflect the true linguistic properties encoded within the language model representations. To address these issues and obtain more robust probing results, recent advancements have introduced innovative techniques, such as the Minimum Description Length (MDL) probing approach proposed by \citet{voita-titov-2020-information}.

In MDL probing, the objective is not solely to assess the accuracy of the shallow classifier but also to measure the effort required to extract the targeted linguistic information from the model representations.
Formally, they establish that a code exists to losslessly compress the labels using Shannon-Huffman code such that $L_p(y_{1,z}|x_{1,z}) = -\sum_{i=1}^z log_2 p(y_i|x_i)$. Note that this is the cross-entropy loss. Furthermore, they define the uniform code length as  $L^{\text{unif}}(y_{i,z}|x_{i,z}) = z log_2(C)$ where $C$ is the number of classes in our task.
 
Given a model $P_\theta(y|x)$ with learnable parameters $\theta$, they choose blocks $1 = n_0 < n_1 < ... < n_s = N$ and encode data by these blocks. The model starts by transmitting the data using the uniform code length for the first chunk. The model is then trained to predict labels $y$ from the data $x$, and also used to predict the labels.
The next block is transmitted using this trained new model. This process continues until the entire dataset is covered. Online code length is calculated as follows: 
\begin{equation}
\begin{aligned}
&L^{\text {online }}\left(y_{1: z} \mid x_{1: z}\right)=z_{1} \log _{2} C \\
&-\sum_{i=1}^{S-1} \log _{2} p_{\theta_{i}}\left(y_{n_{i}+1: n_{i+1}} \mid x_{n_{i}+1: n_{i+1}}\right)
\end{aligned}\end{equation}
Note that this encourages the model to perform well with smaller blocks, as if the model performs well in compressing the data in the block $n_i$, the compression will be increased for the subsequent block $n_{i+1}$.

Having calculated the code lengths, they compare the cross-entropy loss against the uniform code length to find the final compression. 
Formally, compression ($\mathcal{C}$) is defined as the ratio 
$\frac{L^{\text{online}}}{L^{\text{unif}}}$, quantifying how much the model compresses gender information relative to a uniform baseline.

\section{Methodology}
\label{sec:method}

For this study, we focus on gender information as the knowledge property being probed. We will employ MDL probing to evaluate this phenomenon.

\paragraph{Models.} 
Our experiments analyze the representations generated by a diverse range of models. We primarily focus on BERT \cite{devlin-etal-2019-bert}, ALBERT \cite{DBLP:conf/iclr/LanCGGSS20}, and RoBERTa \cite{DBLP:journals/corr/abs-1907-11692}, which are widely used architectures in NLP, and we explore different variations and sizes of these models. Additionally, we examine with a newer model architecture called JINA Embeddings \cite{DBLP:journals/corr/abs-2310-19923}, which is popular in retrieval-augmented generation (RAG) pipelines. This model architecture offers a promising alternative due to the long context size and competitive performance, as claimed by the authors. By comparing these models, we aim to identify common patterns in how they encode gender information and assess their performance in mitigating biases.

\paragraph{Probing Dataset.} 
 We use the Bias in Bios dataset \cite{DBLP:conf/fat/De-ArteagaRWCBC19}, which consists of 396,347 biographies. In this dataset, the gender of each individual is provided as a label alongside their occupation. This allows us to explore how gender information is encoded in language models when analyzing these biographies. In the Bias in Bios dataset, each data point is structured as a triplet $\{X, Y, Y_p\}$, where $X$ represents a biography, $Y$ denotes the true occupation label from one of 28 possible categories, and $Y_p$ indicates the gender of the person featured in the biography. 

\paragraph{Bias Definition and Implications.}
We formally define bias in terms of gender information encoding using the MDL probing framework. Let $f_\theta: X \to Z$ represent a language model with parameters $\theta$ that transforms input text $X$ into latent representations $Z$. Let $f_{\theta_{\text{rand}}}$ be the same model architecture but with randomly initialized weights $\theta_{\text{rand}}$. We denote the compression of gender information from these representations using online code length as $\mathcal{C}_{\theta}$ and $\mathcal{C}_{\theta_{\text{rand}}}$ respectively.

A model $f_\theta$ exhibits gender bias at layer $l$ if the gender information can be extracted with significantly higher compression compared to a randomly initialized model with the same architecture:
\begin{equation}
\mathcal{C}_{\theta^l} - \mathcal{C}_{\theta_{\text{rand}}^l} > \delta
\end{equation}
where $\theta^l$ and $\theta_{\text{rand}}^l$ represent the model parameters at layer $l$ for the trained and randomly initialized models respectively, and $\delta > 0$ is a threshold determining the significance of the difference.

If a model encodes significant gender information, it could use this in decision-making, which is problematic for tasks like Bias in Bios, where we aim to predict occupations without relying on gender. This issue extends to retrieval tasks, such as systems finding resumes for job positions, where gender should not influence results. If retrieval models use gender information, they could reinforce biases that propagate through LLM workflows, leading to unfair outcomes and reinforcing stereotypes. Addressing this bias is essential for creating fairer and more ethical systems.

\begin{figure*}[ht!]
  \includegraphics[width=\linewidth]{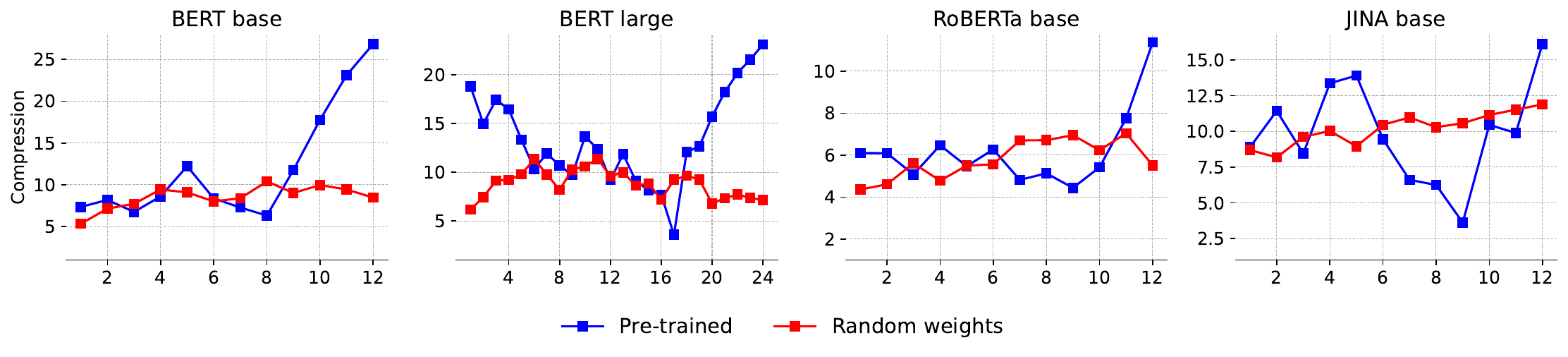}
  \caption {Gender information compression across different layers for various encoder models.}
  \label{fig:compressions}
\end{figure*}

\section{Gender Encoding Analysis}

Building upon the framework outlined in the previous sections,
we conducted our main experiment to investigate whether there is a consistent pattern in how different encoder models encode gender information within their representations. Our primary goal was to determine if various models, despite architectural differences, exhibit similar behaviors in the way they handle gender-related information across their layers.

We experimented with a diverse range of encoder models to ensure the robustness of our findings. The main models discussed in this part are BERT-base, BERT-large, RoBERTa-base, and the base version of JINA Embeddings; however, we also saw similar results with ALBERT and a small version of JINA Embeddings. The results from these models are not included here due to space constraints.

Using the MDL probing method, we measured the amount of gender information that can be compressed from the representations at each layer of these models. Figure \ref{fig:compressions} illustrates the compression rates of gender information across different layers for the selected models. For each layer of the model, we also included a random baseline, which involves calculating compression for each layer of a model initialized with the same architecture but random weights. 
This baseline serves as a control to determine whether the observed compression is due to meaningful encoding of gender information or merely random noise.

Analyzing the results, we observed that models start with varying amounts of encoded gender information in their initial layers: while smaller models, like BERT base, do not exhibit gender information compression in their initial layers, larger models, such as BERT large, show high compression right from the first layer. 

A consistent pattern emerges across all models. Initially, the models seem to reduce the gender information signal within their representations. This reduction continues up to a certain layer, typically close to the final layers. At this critical point, the compression rate of the random baseline representations becomes notably higher than that of the actual model's representations. Beyond this point, the models begin to reconstruct the gender information within their representations. By the final layer, all models demonstrate the highest amount of compression of gender information compared to any other layer. This indicates that, after initially suppressing the gender signal, the models ultimately encode it strongly in their final representations.

This pattern suggests a two-phase process in how encoder models handle gender information: (i) In the early layers, models may abstract away from specific attributes like gender, focusing instead on general linguistic features. (ii) In the later layers, models reintroduce and amplify gender-related information, potentially utilizing it for downstream tasks but also risking the propagation of bias.
These insights underscore the pervasive nature of bias in language models and the need for targeted strategies to mitigate it, particularly in the layers where gender information is reintroduced.

\begin{table*}[t!]
\centering
\resizebox{.65\linewidth}{!}{%
\begin{tabular}{
  l
  l
  S[table-format=2.2]
  S[table-format=2.2]
  S[table-format=2.2]
}
\toprule
\bf{Model}                      & \bf{Technique Name}   & \bf{CrowS-Pairs}      & \bf{StereoSet}    & \bf{DiFair (GNS)} \\ \midrule
\multirow{5}{*}{BERT-base}      & Vanilla               & 58.02                 & 62.02             & 63.91             \\ \cmidrule(l){2-5}
                                & CDA                   & 51.15 \da{6.87}       & 72.98 \uag{10.96} & 86.44 \uag{22.53} \\
                                & Dropout               & 57.25 \da{0.77}       & 66.45 \uag{4.43}  & 68.59 \uag{4.68}  \\
                                & Orthogonal Projection & 53.44 \da{4.58}       & 66.00 \uag{3.98}  & 60.46 \dab{3.45}   \\
                                & ADELE                 & 54.20 \da{3.82}       & 64.76 \uag{2.74}  & 80.21 \uag{16.30} \\ \midrule
\multirow{5}{*}{RoBERTa-base}   & Vanilla               & 54.96                 & 66.50             & 73.38             \\ \cmidrule(l){2-5}
                                & CDA                   & 51.15 \da{3.81}       & 63.59 \dab{2.91}  & 82.58 \uag{9.20}  \\
                                & Dropout               & 53.44 \da{1.52}       & 69.26 \uag{2.76}  & 78.90 \uag{5.52}  \\
                                & Orthogonal Projection & 51.53 \da{3.43}       & 69.19 \uag{2.69}  & 80.27 \uag{6.89}  \\
                                & ADELE                 & 49.62 \da{5.34}       & 65.88 \dab{0.62}  & 70.67 \dab{2.71}   \\ \midrule
\multirow{4}{*}{BERT-large}     & Vanilla               & 55.34                 & 63.99             & 58.70             \\ \cmidrule(l){2-5}
                                & Pretrained CDA         & 53.82 \da{1.52}       & 70.59 \uag{6.60} & 84.26 \uag{25.56} \\
                                & Pretrained Dropout     & 46.56 \da{8.78}       & 54.95 \dab{9.04} & 91.09 \uag{32.39} \\
                                & Post-Hoc CDA          & 56.87 \ua{1.53}       & 69.14 \uag{5.15}  & 84.56 \uag{25.86} \\
                                & Post-Hoc Dropout      & 57.63 \ua{2.29}       & 67.45 \uag{3.46}  & 64.03 \uag{5.33}   \\ \bottomrule
\end{tabular}
}
\caption{Evaluation of debiasing on model weights for three benchmarks. ``Metric Score'' from CrowS-Pairs aims for 50; deviations suggest gender bias. ``ICAT Score'' and ``Gender Neutrality Score'' aim for 100 on StereoSet and DiFair, respectively.}
\label{tab:intrinsic_eval}
\end{table*}

\section{Impact of Bias Mitigation}

Bias mitigation in language models seeks to address both overt biases in model outputs and the subtler, systemic biases embedded within the model's internal representations. Effective techniques should suppress these encoded biases while maintaining model utility. In this section, we investigate the impact of various debiasing methods on compression values, used as a measure of encoded gender information, and evaluate their effectiveness across different experimental setups and models.

\subsection{Experimental Settings}

The experiments assess the performance of four debiasing methods applied to encoder-based language models, including BERT (base and large) and RoBERTa base. We begin by validating the correct implementation of the debiased variations of these models using a series of intrinsic benchmarks, as all debiasing techniques evaluated are intrinsic in nature. Specifically, we employ the CrowS-Pairs \cite{nangia-etal-2020-crows}, StereoSet \cite{nadeem-etal-2021-stereoset}, and DiFair \cite{zakizadeh-etal-2023-difair} benchmarks. The evaluation results for these benchmarks are summarized in Table \ref{tab:intrinsic_eval}. The findings indicate that all debiased models demonstrate effectiveness, with at least two benchmarks showing improved fairness metrics compared to their vanilla counterpart. 

\paragraph{Overview of Debiasing Techniques}
We employed four distinct debiasing strategies to assess the impact of debiasing on model representations. Counterfactual Data Augmentation (CDA; \citealp{zhao-etal-2018-gender}) replaces gendered terms with neutral counterparts and retrains the model on the augmented data, effectively neutralizing biased associations. Adapter-Based Debiasing (ADELE; \citealp{lauscher-etal-2021-sustainable-modular}) uses CDA-augmented data to train modular adapters that reduce bias without retraining the entire model. Dropout applies higher dropout rates during training, hypothesizing that enhanced regularization can reduce encoded biases \cite{webster-measuring}. Finally, Orthogonal Projection \cite{kaneko-bollegala-2021-debiasing} removes gender-related components from intermediate representations through linear projections, offering a lightweight post-hoc solution. Among the described bias mitigation techniques, ADELE and Orthogonal Projection are inherently post-hoc methods. Conversely, CDA and Dropout may be implemented at any stage, either during the post-hoc phase or from the onset of training.

\paragraph{Debiasing Effectiveness}
Based on our experiments in the previous section, gender-related information predominantly concentrates in the initial and final layers of the examined models. Given our formal definition of gender bias, we can precisely define the effectiveness of a debiasing method. Let $f_{\theta_{\text{debias}}}$ represent a model after applying a debiasing technique, with $\theta_{\text{debias}}$ denoting its parameters, and $f_{\theta}$ the original vanilla model with parameters $\theta$. An ideal debiasing method is considered effective if it satisfies:
\begin{equation}
\begin{split}
\mathcal{C}_{\theta_{\text{debias}}^l} &\leq \min ( \mathcal{C}_{\theta^l} , \mathcal{C}_{\theta_{\text{rand}}^l} + \delta )
\end{split}
\end{equation}
where $\theta_{\text{debias}}^l$, $\theta^l$, and $\theta_{\text{rand}}^l$ represent the parameters at layer $l$ for the debiased model, vanilla model, and randomly initialized model respectively, $L$ denotes the total number of layers, and $\delta \geq 0$ is our bias significance threshold.

In simple terms, a debiasing method is effective if, across all layers, it reduces the compression of gender information below both the vanilla model and the threshold established by the random baseline. This indicates successful elimination of the gender signal from the representations throughout the entire model architecture. Conversely, if a method fails to satisfy this criterion at any layer, it indicates that the debiasing approach is ineffective or even counterproductive in terms of compression.

\begin{figure*}[ht!]
  \includegraphics[width=\linewidth]{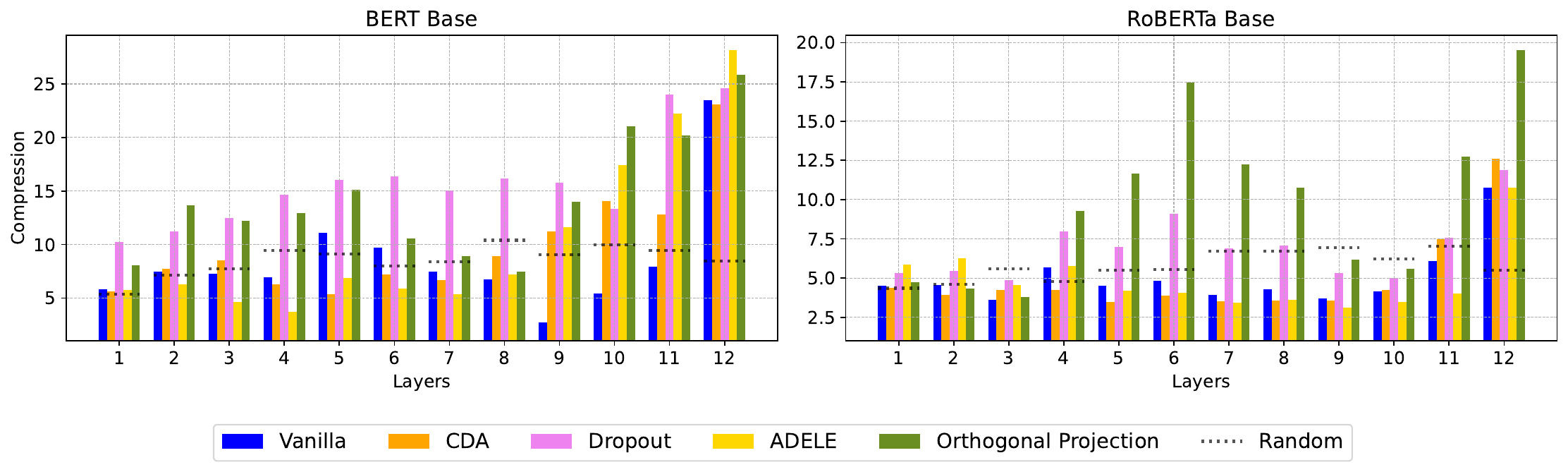}
  \caption {Effect of various bias mitigation procedures on gender information compression across different layers of base models.}
  \label{fig:debias_compressions}
\end{figure*}

\begin{figure*}[ht!]
  \includegraphics[width=\linewidth]{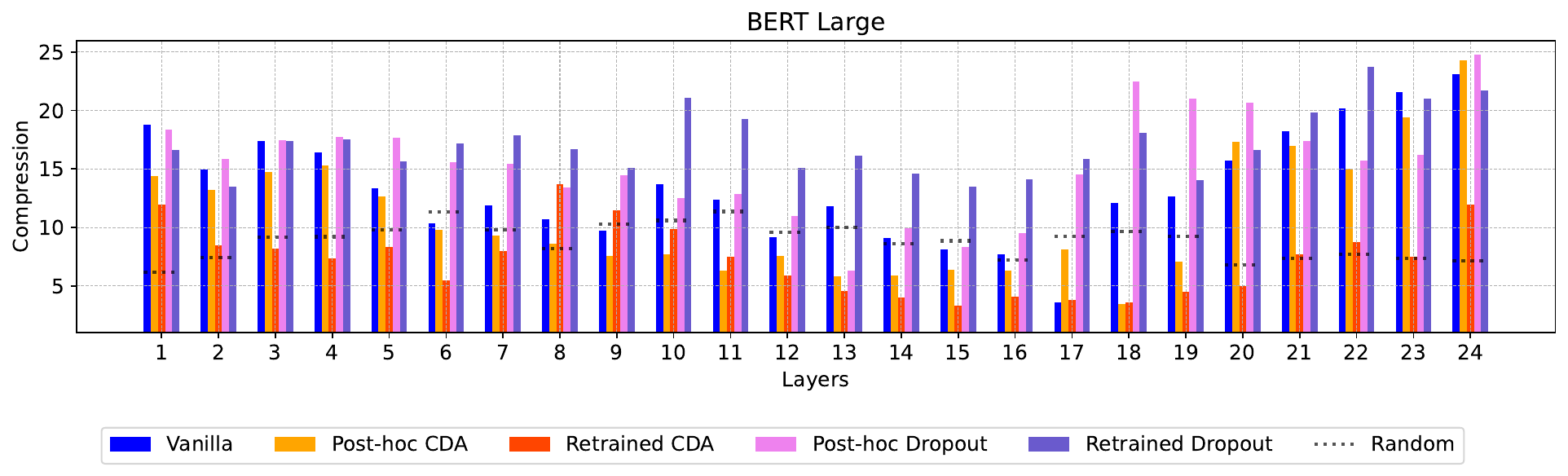}
  \caption {Effect of various bias mitigation procedures on gender information compression across different layers of BERT-Large.}
  \label{fig:compressions_bert-large}
\end{figure*}

\subsection{Results and Analysis}

The results of these experiments are presented in Figures \ref{fig:compressions} and \ref{fig:compressions_bert-large}. Our analysis reveals that, with the exception of training-time CDA, the remaining methods were ineffective in reducing bias in the models. Some methods, such as ADELE and training-time Dropout, show mixed results, suggesting that their effectiveness may be influenced by factors such as model architecture and training parameters. In the following discussion, we will elaborate on these observations in detail.

\paragraph{Layer-Wise Trends in Compression}  
Compression values exhibited a consistent pattern across all models. In the lower layers, gender information was minimally compressible, suggesting that these layers encode relatively little bias. However, in the final layers, compression values increased sharply, indicating that gender information becomes more concentrated and accessible as representations become more abstract.

\paragraph{Impact of Training-Time Debiasing}  
Training-time CDA on BERT-large demonstrated the most substantial reduction in final-layer compression. The compression value in the final layer decreased from 23.08 in the vanilla model to 11.98 after retraining with CDA, confirming its effectiveness in suppressing gender information throughout the model. Similarly, training-time dropout resulted in a lower final-layer compression compared to the vanilla model, though its effect was less pronounced than CDA.

\paragraph{Effectiveness of Post-Hoc Methods}  
Post-hoc CDA and dropout, applied across all models, were generally less effective in mitigating gender encoding. In BERT-large, post-hoc CDA failed to achieve the same level of suppression as training-time CDA, resulting in a final-layer compression of 20.34. Dropout exhibited inconsistent behavior across models; in some cases, it preserved or even amplified gender information. For instance, in BERT-base, the final-layer compression increased from 23.47 (vanilla) to 24.63 with post-hoc dropout, indicating that this method does not reliably suppress bias.

\paragraph{Comparison Across Model Architectures}  
RoBERTa-base consistently displayed lower compression values than BERT-based models, suggesting that its architecture inherently encodes less gender-related information. This observation aligns with its performance on intrinsic bias benchmarks, where it demonstrated reduced sensitivity to gendered associations. 
Comparing BERT-base and BERT-large also indicates that larger models tend to store more gender information in their representations, which also aligns with the results obtained from the intrinsic bias benchmarks. This suggests that as model capacity increases, so does its ability to encode and retain gendered associations, reinforcing the need for targeted mitigation strategies in larger models.

While all debiasing methods contributed to reducing gender encoding to some extent, none completely eliminated it across all layers. Training-time CDA proved the most effective strategy, whereas post-hoc methods showed limited success, particularly in mitigating gender encoding in the final layers. These findings indicate that bias is deeply ingrained in model representations and that effective mitigation requires intervention during training rather than post-hoc adjustments.

For practical applications where reducing gender encoding is a priority, retraining with targeted debiasing objectives remains the most reliable approach. Future work could explore hybrid strategies that combine training-time and post-hoc techniques to enhance bias suppression without requiring full retraining.

\begin{figure*}[ht!]
  \includegraphics[width=\linewidth]{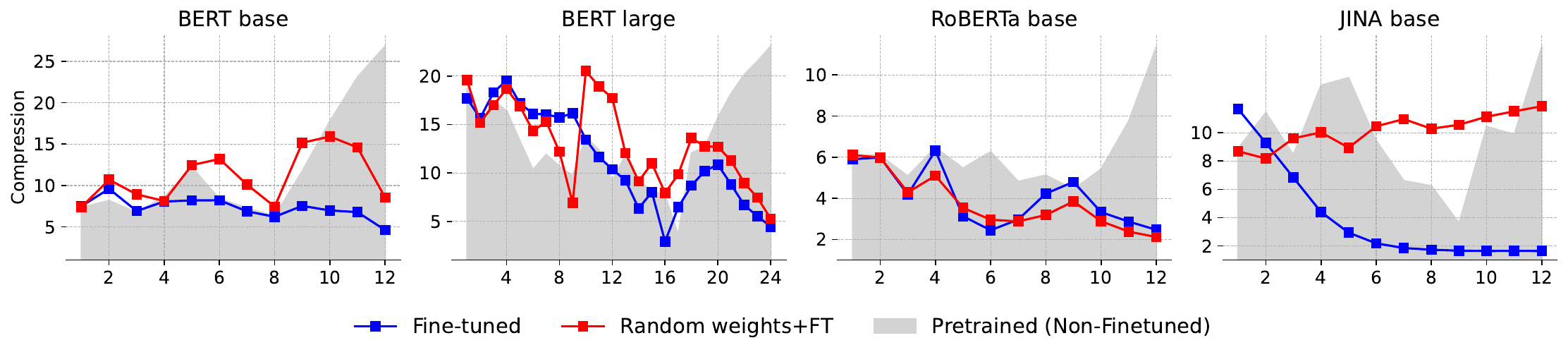}
  \caption {Gender information compression across different layers for the fine-tuned encoder models. The pretrained compression values correspond to the blue line shown in Figure \ref{fig:compressions}.
}
  \label{fig:finetuned_compressions}
\end{figure*}

\section{Impact of Fine-tuning}

While encoder models are widely used in retrieval systems, their representations are typically fine-tuned for downstream tasks such as classification. Understanding how this process influences gender bias encoded in model representations is critical, as fine-tuning may alter or amplify existing biases. In this section, we investigate how fine-tuning affects gender-related information stored in model layers and evaluate its implications for bias mitigation.

\subsection{Experimental Settings}
We fine-tuned three encoder models -- BERT-base, BERT-large, and RoBERTa-base -- on the BiosBias dataset. The task involves predicting an individual’s occupation from their biography, framed as a 28-class classification problem. Models were trained for 5 epochs using a learning rate of $2\times10^{-5}$. To isolate the impact of fine-tuning, we compared the fine-tuned models against two baselines: (i) their original pretrained versions and (ii) "randomized" counterparts initialized with untrained weights but fine-tuned on the same task. Layer-wise MDL probing was applied to all models to measure gender information compression before and after fine-tuning.

\subsection{Results and Analysis}

The experimental results, presented in Figure \ref{fig:finetuned_compressions}, reveal several noteworthy patterns in how fine-tuning affects gender information encoding.

\paragraph{Reduced Gender Information}
Fine-tuning consistently led to a substantial reduction in gender information compression across all models. This reduction was particularly pronounced in the final layers, where the original models had shown the highest concentration of gender information.

\paragraph{Below-Random Compression}
In many cases, the compression values of fine-tuned models fell below those of their random baselines. Notably, even the random baselines of fine-tuned models showed lower compression compared to their pretrained counterparts. This suggests that task-specific fine-tuning may actively suppress the encoding of gender information in favor of task-relevant features.

\paragraph{Shift in Representational Focus}
The dramatic reduction in gender information compression indicates that fine-tuning redirects the model's internal representations toward task-specific features and away from demographic attributes like gender. This finding suggests that much of the bias observed in fine-tuned models may originate from the classification head rather than from biases encoded in the underlying representations.

These findings carry significant implications for bias mitigation in language models. The observation that fine-tuning naturally reduces encoded gender information while potentially concentrating bias in the classification layer explains the limited impact of intrinsic debiasing methods on extrinsic bias metrics \cite{orgad-etal-2022-gender, cao-etal-2022-intrinsic}. While task-specific fine-tuning may serve as an implicit form of representation-level bias mitigation, our results suggest that future debiasing efforts should focus more on the classification components added during fine-tuning rather than the encoder representations alone.

\section{Conclusions}

Our analysis reveals that pretrained language models follow a consistent pattern of gender encoding: early layers suppress gender signals, while later layers amplify them, embedding bias deeply into abstract representations. Current debiasing techniques, particularly post-hoc interventions, show limited efficacy in altering these internal mechanisms. Task-specific fine-tuning reduces encoded gender information but risks concentrating residual bias in downstream classifiers, underscoring the need for holistic mitigation strategies that target both representations and decision layers.
Collectively, these findings challenge conventional debiasing paradigms, advocating for proactive integration of fairness objectives during pretraining and architecture-aware interventions targeting bias propagation pathways.

\section*{Broader Impacts}

Our results have significant implications for the design and deployment of language models. First, they underscore the inadequacy of post-hoc debiasing methods, urging researchers to integrate fairness objectives directly into pretraining. Second, the localization of bias in later layers suggests targeted interventions, such as modifying specific layers or attention heads, could offer efficient mitigation pathways. Finally, practitioners must recognize that reducing bias in representations does not guarantee fairness in downstream applications; rigorous evaluation of classifiers and datasets remains essential. These insights advocate for a paradigm shift toward inherently fair model architectures and training frameworks.

\section*{Limitations}

While this work provides critical insights, several limitations warrant consideration. First, our analysis focuses on gender bias in English-language biographies, leaving broader sociocultural and intersectional biases unexplored. Second, the study centers on encoder-based models; future work should validate findings in decoder-based architectures and multimodal systems. Lastly, the interplay between task-specific fine-tuning and bias propagation requires deeper exploration across diverse applications. Addressing these gaps will advance our understanding of bias dynamics and mitigation in increasingly complex language technologies.

\bibliography{bib/anthology,bib/others}

\end{document}